# Zero-Shot Multi-Disease Labeling of Chest, Abdomen, and Pelvis CT Reports Using Open-Weight Large Language Models: The Effect of Labeling Conventions

Michael E. Garcia-Alcoser MS [1,2], Mobina GhojoghNejad MD [1], Fakrul Islam Tushar PhD [3], David Kim MS [1], Kyle J. Lafata PhD [1,2], Geoffrey D. Rubin MD [4], and Joseph Y. Lo PhD [1,2]

1. Center for Virtual Imaging Trials, Carl E. Ravin Advanced Imaging Laboratories, Department of Radiology, Duke University School of Medicine, Durham, NC, USA
2. Medical Physics Graduate Program, Duke University, Durham, NC, USA
3. Department of Radiology & Imaging Sciences, University of Arizona College of Medicine, Tucson, AZ, USA
4. Department of Medical Imaging, University of Arizona College of Medicine, Tucson, AZ, USA

**Abbreviations:** CT: computed tomography, LLM: large language model, CAP: chest, abdomen, and pelvis, RBA: rule-based algorithm, BERT: bidirectional encoder representations from transformers

**Summary Statement: Zero-shot prompting of lightweight open-weight large language models outperformed rule-based and BERT-based labeling of chest, abdomen, and pelvis CT reports, with most residual disagreement reflecting labeling conventions rather than model error.**

**Key Points:**

- Among five lightweight open-weight large language models labeling 15 disease and normal classes across three organ systems, Gemma-3 27B achieved the highest macro-averaged F1 score (0.82 [95% CI: 0.80, 0.83]), compared with 0.66 [95% CI: 0.64, 0.68] for a fine-tuned RadBERT model and 0.64 [95% CI: 0.62, 0.66] for a rule-based algorithm.
- Labels denoting a normal organ system performed worst across all methods, averaging a macro-averaged F1 score of 0.63 [95% CI: 0.61, 0.65], and hierarchical clustering of Cohen kappa values, separated these labels from the disease labels at the first split, indicating that asserting the absence of disease was more difficult than detecting its presence.
- Relabeling reports by whether a finding was described, rather than whether it was clinically actionable, improved atelectasis labeling for all five large language models (largest ΔF1, 0.32 [95% CI: 0.26, 0.40] for MedGemma 27B) but produced no improvement for the rule-based algorithm or RadBERT, whose rules already excluded gravity-dependent atelectasis.

## Abstract

**Purpose**: To compare five lightweight open-weight large language models (LLMs) with a rule-based algorithm (RBA) and fine-tuned RadBERT for zero-shot labeling of chest-abdomen-pelvis (CAP) CT reports, and to examine how labeling conventions affect measured performance.

**Materials and Methods**: In this retrospective study, 40,833 CAP CT reports from 29,540 patients examined between 2012 and 2017 were analyzed; age and sex were unavailable. Five LLMs were prompted zero-shot to assign 15 labels across three organ systems and compared with an RBA and fine-tuned RadBERT. Inter-model agreement was assessed with Cohen kappa (κ) on 12,197 held-out reports. Macro-averaged F1 was computed against 1,789 radiologist-supervised annotations, the same annotations simplified to disregard clinical actionability, and the CT-RATE dataset. Nonoverlapping bootstrapped 95% CIs indicated relevant differences.

**Results**: MedGemma 27B and MedGemma-1.5 4B showed the highest median agreement (κ = 0.90). Against manual annotations, Gemma-3 27B achieved the highest macro-averaged F1 (0.82 [95% CI: 0.80, 0.83]) versus 0.66 for RadBERT and 0.64 for the RBA; a majority-vote ensemble scored 0.84. Scores were lowest averaged across models for subjective classes, kidney lesion (0.44) and atelectasis (0.67). Relabeling raised F1 for all models on kidney lesion, but for atelectasis only for the LLMs; the RBA and RadBERT declined. F1 against CT-RATE exceeded that against manual annotations for all models, reflecting its more literal convention.

**Conclusion:** Lightweight open-weight LLMs outperformed rule-based and fine-tuned BERT labeling of CAP CT reports with zero-shot prompting. Models and annotators disagreed largely because they applied different labeling criteria.

# Introduction

Computed tomography (CT) images contain rich diagnostic information that underpins numerous deep-learning applications, but tasks such as disease classification and detection require large labeled datasets for training and evaluation (1–3). Manual annotation at this scale is time-consuming and subject to observer variability, and the labels required vary with the intended task.

Radiology reports generated during routine care can instead supply labels for the corresponding images. Progress in extracting findings from free text using rule-based methods, bidirectional encoder representations from transformers (BERT), and large language models (LLMs) has centered on chest radiograph (CXR) reports (4–7). Locally run open-weight LLMs have outperformed rule-based and BERT-based labelers (8–10) and approached closed-weight models such as GPT-4 and GPT-4o, though that gap depends on prompting strategy (8,9). The resulting labels suffice for training downstream vision models (11,12), and richer schemes capturing probability, location, and severity have begun to supplement binary labels (12). Reported performance, however, shifts substantially with the reference standard (13).

CT reports are longer and describe three-dimensional anatomy across multiple organs, encompassing a wider range of pathologies. Work applying LLMs to CT reports has largely been confined to one anatomical region: Wihl et al. extracted ten findings from stroke CT reports, showing that explicit annotation guidelines in the prompt improved accuracy (14), while LEAVS labeled presence and urgency across nine abdominal organs and trained a CT vision model on those labels (15). Both relied on a large closed-weight model or a purpose-built prompting pipeline, leaving open whether lightweight open-weight models, prompted zero-shot, generalize across the anatomically distinct regions of a single chest-abdomen-pelvis (CAP) examination. For CAP reports, a rule-based algorithm (RBA) has been developed and its labels used as weak supervision for organ-specific image classifiers (16,17); preliminary work comparing one open-weight LLM with that RBA and a BiLSTM network found differing conventions for organ normality (18), but used a single model and pseudo-label reference standards, leaving unquantified how much disagreement reflects labeling convention rather than model error.

We assessed five lightweight, open-weight LLMs alongside two established approaches, the RBA and a fine-tuned RadBERT model, for multi-disease annotation of CAP CT reports across three organ systems: kidneys/ureters, liver/gallbladder, and lungs/pleura. Using a previously described institutional CAP CT report dataset (16,17) and the publicly available CT-RATE dataset, we characterized the performance of each method and examined how labeling conventions embedded in a reference standard affect measured performance.

# Materials and Methods

This exploratory retrospective, HIPAA-compliant study was approved by the local Institutional Review Board; the requirement for informed consent was waived.

## Rule-Based Algorithm

We used a previously developed RBA (16) that applies regular-expression rules to produce binary labels for 15 classes. In that work, labels were selected by the prevalence of organ-disease keywords identified using term frequency–inverse document frequency analysis (19), and three organ systems were targeted to vary in location, appearance, and disease manifestation: kidneys/ureters, liver/gallbladder, and lungs/pleura, each contributing four disease labels and one normal label (Table 1). The RBA is applied to the findings section only, so that labels reflect exam-specific findings, and evaluates each sentence for disease-specific or normal descriptors. An organ system is labeled normal only if all four diseases are excluded and a normal descriptor is present; if the organ is not mentioned, all associated labels are treated as negative.

## Dataset

We used a subset of the CAP report corpus described in (16, 17). This subset comprised 40,833 deidentified CAP CT reports from 29,540 patients examined between 2012 and 2017. Patient age and sex were not retained in the deidentified dataset and were therefore unavailable for this analysis. Reports were partitioned by patient into a 70% development set and a 30% test set, the former reserved for fine-tuning RadBERT. Stratified sampling based on RBA pseudo-label frequency made the test set representative of the full dataset, yielding 12,197 reports from 8,854 patients.

## Models

To preserve patient privacy, only open-weight LLMs deployable on local hardware were evaluated. Five lightweight models from two families were assessed: Llama-3.1-8B Instruct, Llama-UltraMedical (Llama-UM, a medically fine-tuned Llama-3.1-8B), Gemma-3-27B, MedGemma-27B-text-it, and MedGemma-1.5-4B-it (20–23). For comparison with a supervised approach, RadBERT-RoBERTa-4m (24) was fine-tuned on the findings sections of the 28,636 development reports using their RBA pseudo-labels.

## Classification Experiments

Experiments were conducted between October 2024 and January 2026. Each of the 12,197 test reports was processed with the transformers package using zero-shot prompts (Figure 1), with one generation per report. Models returned a JavaScript Object Notation (JSON) dictionary containing a present/not-present decision and a supporting sentence extraction for each of the 15 labels (8,9). A majority-voting ensemble combining the RBA, Llama-3.1-8B, and Gemma-3-27B was also evaluated (25). Code for data partitioning, model inference, and statistical analysis, along with the prompts used for all models, is publicly available at (redacted).

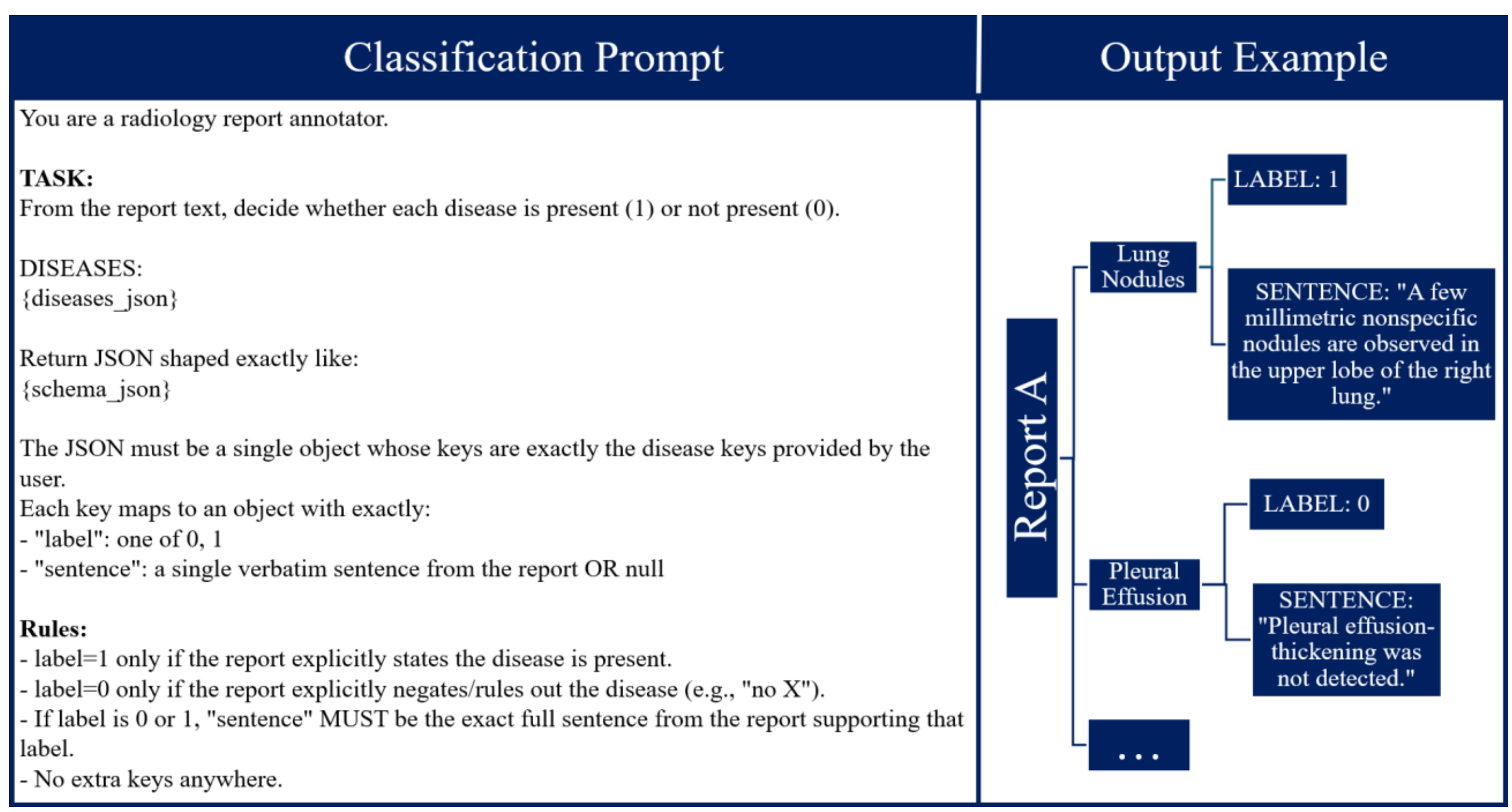

*Figure 1: System prompt to instruct LLMs in multi-label annotation experiments (left) and an output example for a text report (right).*

## Internal Manual Dataset

To assess performance against a reference standard, we manually labeled a subset of the 12,197 CAP test reports. Labeling criteria were defined in advance by a board-certified radiologist with 32 years of experience (redacted). A medical physics student (redacted) and a general physician (redacted) applied these criteria to the reports, referring cases they could not confidently resolve to the radiologist for a final decision. Originally, we sampled reports based on model agreement patterns between RBA, Llama-3.1-8B, and Llama-UM: (1) categorize reports into eight possible agreement/disagreement combinations, (2) randomly sample from each category, and (3) repeat for all classes. Because most reports (81%) showed full agreement, this approach enriched cases where model outputs diverged (19%), yielding 1,564 reports from 1,482 patients. Since enrichment makes F1 scores on this set pessimistic relative to the full test set, an additional 225 reports were randomly sampled to provide an unenriched comparison for models introduced later. The final manual set comprised 1,789 reports from 1,688 patients.

Annotations reflected clinical actionability: findings judged transient or indeterminate, specifically gravity-dependent atelectasis and lesions described as too small to characterize, were labeled negative. Reports whose wording could reasonably support either binary label were additionally categorized as subjective.

## Simplified Manual Dataset

To isolate the effect of these criteria, a simplified reference standard was created in which labels were assigned solely on whether a finding was described. ΔF1 was defined as the F1 score against the simplified labels minus that against the original annotations, so positive values indicate higher performance under the simplified convention.

**Table 1: Class Prevalence for Each Dataset Partition**

| Classes | Development Counts | Test Counts | Manual Counts | Manual Random Counts | Total Manual Counts |
|---|---|---|---|---|---|
| Kidney Stone | 1589 | 640 | 48 | 14 | 62 |
| Kidney Atrophy | 751 | 319 | 68 | 10 | 78 |
| Kidney Lesion | 6059 | 2590 | 25 | 29 | 54 |
| Kidney Cyst | 4652 | 1900 | 82 | 39 | 121 |
| Normal Kidney | 1926 | 824 | 50 | 95 | 145 |
| Gallstone | 2381 | 1029 | 69 | 29 | 98 |
| Liver Lesion | 9108 | 3872 | 53 | 56 | 109 |
| Biliary Dilation | 1708 | 725 | 52 | 12 | 64 |
| Fatty Liver | 1829 | 799 | 63 | 26 | 89 |
| Normal Liver | 2204 | 955 | 48 | 100 | 148 |
| Atelectasis | 7114 | 3027 | 36 | 40 | 76 |
| Lung Nodule | 12769 | 5545 | 61 | 119 | 180 |
| Emphysema | 2772 | 1191 | 40 | 23 | 63 |
| Pleural Effusion | 3879 | 1610 | 38 | 29 | 67 |
| Normal Lung | 787 | 357 | 75 | 48 | 123 |

*Table 1: The prevalence of RBA pseudo labels that are partitioned into development and test sets. The positive counts for each class are also shown for the manual annotated sets (Manual Counts=1,564 reports, Manual Random Counts= 225 reports, Total Manual Counts = 1,789 reports).*

## External CT-RATE Dataset

This publicly available dataset includes 25,692 non-contrast chest CT scans from 21,304 patients. Only 21,304 reports (one report per patient) were used to evaluate model performance. For direct comparison to our dataset, analysis was restricted to the atelectasis, lung nodule, emphysema, and pleural effusion classes (26).

## Statistical Analysis

Inter-model agreement on the 12,197 test reports was assessed using pairwise Cohen kappa (κ) for each label (9), interpreted according to Landis and Koch (27). Classes were hierarchically clustered on their pairwise kappa values across all model comparisons, using median linkage with Euclidean distance. Macro-averaged F1 scores were computed against the manual, simplified, and CT-RATE reference standards, with 95% CIs from 1,000 bootstrap resamples. As this was an exploratory study, nonoverlapping CIs were interpreted as relevant differences; formal significance testing, which would require correction for multiple comparisons, was not performed. Analyses used Python 3.11.5 and scikit-learn 1.5.0 (28).

# Results

## Inter-model agreement

Among the 21 model pairs evaluated on the 12,197 CAP test reports, six showed median "almost perfect" agreement. MedGemma 27B and MedGemma-1.5 4B agreed most closely (median κ = 0.90 [IQR: 0.65, 0.92]), followed by Llama-3.1 8B and Gemma-3 27B (0.87 [IQR: 0.73, 0.90]). The RBA and RadBERT also showed almost perfect agreement (0.85 [IQR: 0.83, 0.91]). The remaining pairs showed median substantial agreement, with the RBA and Llama-UM lowest (0.64 [IQR: 0.44, 0.70]). Llama-3.1 8B and Llama-UM agreed less closely than other within-family pairs (0.76 [IQR: 0.55, 0.83]). Across all the model comparisons, the emphysema class had the highest κ median of 0.94 [IQR: 0.91, 0.96] and normal kidney had the lowest κ median of 0.27 [IQR: 0.13, 0.37]. Hierarchical clustering of the class-level kappa values separated the labels into two groups at the first bifurcation, corresponding to the disease and normal classes; the normal classes showed the lowest agreement across all model comparisons (Figure 2).

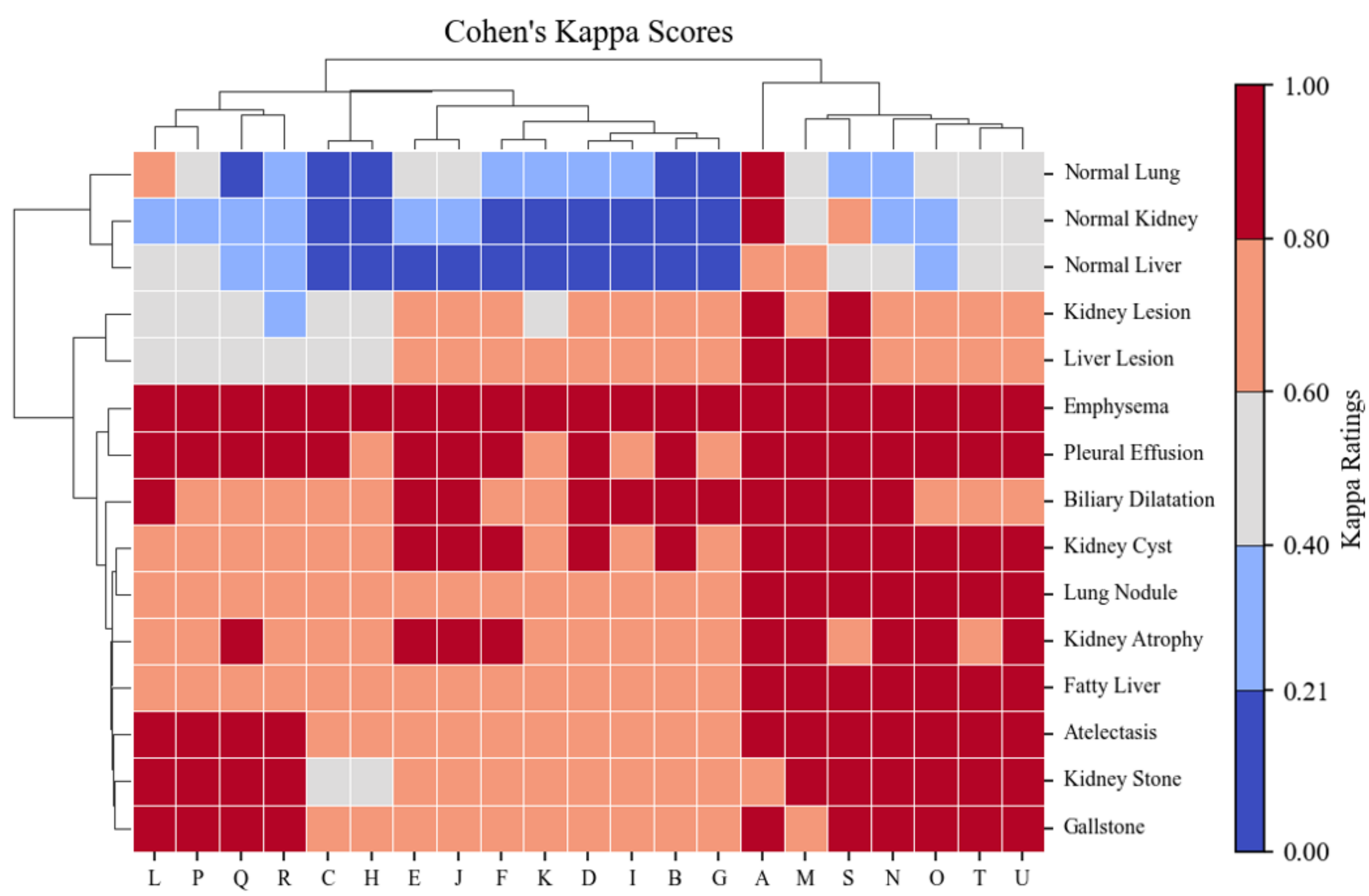

| **A: RBA – RadBERT** | H: RadBERT – Llama-UM | **O: Llama – MedGemma-1.5 4B** |
|---|---|---|
| B - RBA – Llama | I: RadBERT – Gemma-3 | P: Llama-UM – Gemma-3 |
| C: RBA – Llama-UM | J: RadBERT – MedGemma 27B | Q: Llama-UM - MedGemma 27B |
| D: RBA – Gemma-3 | K: RadBERT – MedGemma-1.5 4B | R: Llama-UM – MedGemma-1.5 4B |
| E: RBA – MedGemma 27B | L: Llama – Llama-UM | **S: Gemma-3 – MedGemma 27B** |
| F: RBA – MedGemma-1.5 4B | **M: Llama – Gemma-3** | **T: Gemma-3 – MedGemma-1.5 4B** |
| G: RadBERT – Llama | **N: Llama – MedGemma 27B** | **U: MedGemma 27B – MedGemma-1.5 4B** |

*Figure 2: This table presents a clustered heatmap of the Cohen κ statistics for each pairwise comparison of predictions on 12,197 CAP CT reports. The dark blue represents the lowest inter-rater agreement, and the dark red represents the highest inter-rater agreement. The bolded comparisons A, M, S, N, O, T and U, achieved the "almost perfect" agreement category.*

## Model Evaluation

Using the internal manual test set as the reference standard, Gemma-3 27B achieved the highest macro F1 score at 0.82 [95% CI: 0.80, 0.83]. Llama-3.1 8B and MedGemma 27B both tied with scores of 0.79 [95% CI: 0.77, 0.80]. MedGemma-1.5 4B resulted in a macro F1 score of 0.78 [95% CI: 0.76, 0.80]. The other three models, Llama-UM, RadBERT, and RBA, all demonstrated lower macro F1 scores of 0.66 [95% CI: 0.64, 0.68], 0.66 [95% CI: 0.64, 0.68], and 0.64 [95% CI: 0.62, 0.66], respectively. A majority vote ensemble, composed of the RBA, Llama-3.1 8B, and Gemma-3 27B, achieved a slightly higher macro F1 score than the best individual model of Gemma-3 27B alone, at 0.84 [95% CI: 0.83, 0.85]. For visualization purposes, the F1 scores for the RBA, Llama-3.1 8B, Gemma-3 27B, and the majority vote ensemble can be seen in Figure 3. The F1 scores for all classes and models can be found in Table 2.

An organ-level performance analysis revealed variation in F1 scores between the models. For the kidney/ureters system, MedGemma 27B achieved the highest F1 of 0.97 [95% CI: 0.93, 0.99] for kidney stone, while the RBA and RadBERT had the lowest score of 0.36 [95% CI: 0.26, 0.45] for kidney lesion. In the liver/gallbladder system, MedGemma-1.5 4B reached the highest F1 of 0.97 [95% CI: 0.95, 0.99] for gallstone, while the RBA attained an F1 of 0.45 [95% CI: 0.36, 0.53] for normal liver. Lastly, in the lungs/pleura system, both MedGemma 27B and MedGemma-1.5 4B recorded the highest F1 of 0.97 [95% CI: 0.94, 0.99] and 0.97 [95% CI: 0.93, 0.99] respectively for emphysema. MedGemma 27B had the lowest F1 of 0.46 [95% CI: 0.36, 0.55] for normal lung. The gallstone and emphysema classes achieved the highest average performance across all models 0.85 [95% CI: 0.83, 0.87]. In contrast, kidney lesion and atelectasis had the lowest F1 scores among disease classes 0.44 [95% CI: 0.42, 0.46] and 0.67 [95%

CI: 0.65, 0.69], respectively. The normal labels for the kidney, liver, and lung each had an average F1 score of 0.63 [95% CI: 0.61, 0.65] across all models.

**Table 2: Manual Dataset F1 Scores**

| Classes | RBA | RadBERT | Llama-3.1 8B | Llama-UM | Gemma-3 27B | MedGemma 27B | MedGemma-1.5 4B | Majority |
|---|---|---|---|---|---|---|---|---|
| **Kidney Stone** | 0.60 [0.50,0.70] | 0.69 [0.57,0.79] | 0.88 [0.82,0.93] | 0.79 [0.70,0.87] | 0.95 [0.91,0.99] | 0.97 [0.93,0.99] | 0.94 [0.90,0.98] | 0.95 [0.91,0.98] |
| **Kidney Atrophy** | 0.59 [0.48,0.69] | 0.61 [0.50,0.71] | 0.82 [0.75,0.88] | 0.72 [0.63,0.81] | 0.88 [0.83,0.93] | 0.80 [0.72,0.87] | 0.75 [0.66,0.83] | 0.85 [0.77,0.91] |
| **Kidney Lesion** | 0.36 [0.26,0.45] | 0.36 [0.26,0.45] | 0.47 [0.38,0.55] | 0.43 [0.31,0.55] | 0.51 [0.41,0.60] | 0.54 [0.44,0.63] | 0.42 [0.34,0.50] | 0.50 [0.41,0.59] |
| **Kidney Cyst** | 0.82 [0.75,0.87] | 0.82 [0.77,0.87] | 0.85 [0.80,0.90] | 0.52 [0.42,0.61] | 0.94 [0.90,0.96] | 0.94 [0.91,0.97] | 0.92 [0.89,0.96] | 0.95 [0.93,0.98] |
| **Normal Kidney** | 0.39 [0.30,0.47] | 0.41 [0.32,0.49] | 0.87 [0.82,0.90] | 0.60 [0.52,0.66] | 0.81 [0.75,0.85] | 0.65 [0.58,0.71] | 0.69 [0.63,0.75] | 0.84 [0.79,0.88] |
| **Gallstone** | 0.74 [0.67,0.81] | 0.75 [0.68,0.82] | 0.77 [0.70,0.83] | 0.89 [0.84,0.93] | 0.90 [0.86,0.95] | 0.92 [0.88,0.96] | 0.97 [0.95,0.99] | 0.93 [0.89,0.96] |
| **Liver Lesion** | 0.64 [0.56,0.71] | 0.72 [0.65,0.79] | 0.76 [0.70,0.82] | 0.59 [0.49,0.68] | 0.78 [0.72,0.84] | 0.74 [0.67,0.80] | 0.73 [0.67,0.79] | 0.79 [0.73,0.84] |
| **Biliary Dilation** | 0.70 [0.60,0.78] | 0.80 [0.70,0.87] | 0.80 [0.72,0.87] | 0.69 [0.58,0.78] | 0.87 [0.81,0.93] | 0.79 [0.69,0.87] | 0.74 [0.65,0.82] | 0.91 [0.86,0.96] |
| **Fatty Liver** | 0.65 [0.55,0.74] | 0.59 [0.49,0.69] | 0.93 [0.89,0.97] | 0.70 [0.62,0.78] | 0.93 [0.89,0.96] | 0.85 [0.79,0.91] | 0.90 [0.85,0.95] | 0.93 [0.89,0.96] |
| **Normal Liver** | 0.45 [0.36,0.53] | 0.47 [0.39,0.55] | 0.82 [0.78,0.87] | 0.65 [0.59,0.71] | 0.73 [0.67,0.77] | 0.70 [0.64,0.76] | 0.57 [0.50,0.64] | 0.78 [0.73,0.83] |
| **Atelectasis** | 0.82 [0.75,0.88] | 0.81 [0.74,0.87] | 0.61 [0.53,0.68] | 0.51 [0.42,0.60] | 0.65 [0.58,0.72] | 0.64 [0.56,0.71] | 0.66 [0.59,0.73] | 0.68 [0.61,0.75] |
| **Lung Nodule** | 0.81 [0.76,0.85] | 0.81 [0.77,0.86] | 0.89 [0.85,0.92] | 0.71 [0.65,0.76] | 0.88 [0.84,0.91] | 0.91 [0.89,0.94] | 0.89 [0.85,0.92] | 0.92 [0.89,0.94] |
| **Emphysema** | 0.71 [0.63,0.79] | 0.70 [0.61,0.79] | 0.83 [0.76,0.89] | 0.79 [0.71,0.86] | 0.94 [0.90,0.98] | 0.97 [0.94,0.99] | 0.97 [0.93,0.99] | 0.96 [0.93,0.99] |
| **Pleural Effusion** | 0.76 [0.68,0.83] | 0.74 [0.65,0.82] | 0.81 [0.74,0.88] | 0.67 [0.58,0.76] | 0.80 [0.72,0.86] | 0.90 [0.84,0.95] | 0.87 [0.81,0.93] | 0.86 [0.80,0.92] |
| **Normal Lung** | 0.65 [0.56,0.72] | 0.60 [0.52,0.68] | 0.74 [0.68,0.80] | 0.62 [0.54,0.69] | 0.71 [0.64,0.78] | 0.46 [0.36,0.55] | 0.66 [0.58,0.73] | 0.75 [0.68,0.81] |
| **Macro Average** | 0.64 [0.62,0.66] | 0.66 [0.64,0.68] | 0.79 [0.77,0.80] | 0.66 [0.64,0.68] | 0.82 [0.80,0.83] | 0.79 [0.77,0.80] | 0.78 [0.76,0.80] | 0.84 [0.83,0.85] |

*Table 2: F1 Scores for each model, against manual dataset. The numbers in the bracket represent the 95% confidence interval.*

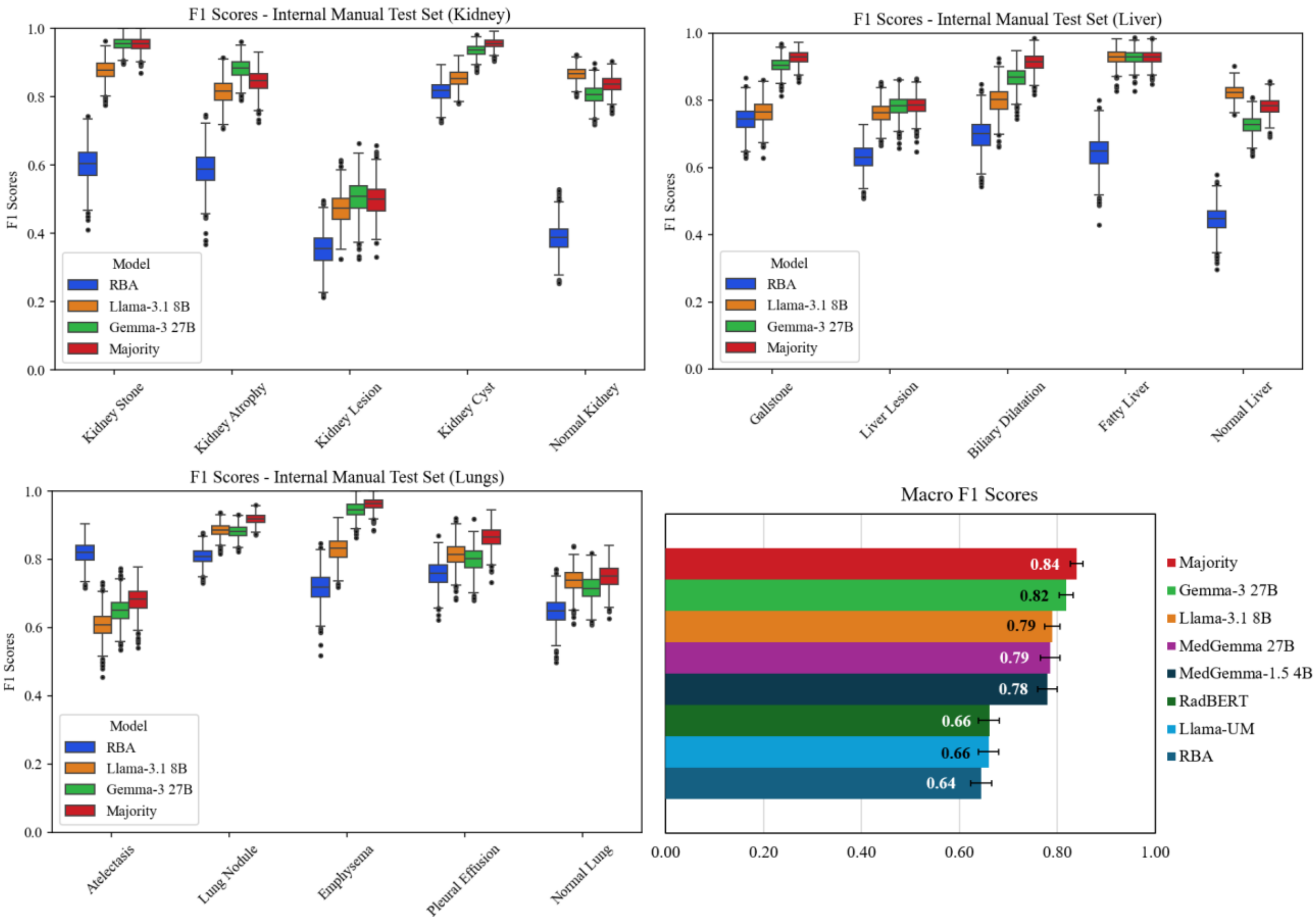

*Figure 3: F1 scores for each of the 15 disease labels across four models: RBA, Llama-3.1 8B, Gemma-3 27B, and the majority vote ensemble. Sub-figures show results for labels from the kidneys/ureters, liver/gallbladder, and lungs/pleura, and bar chart in the bottom-right corner displays the macro F1 scores for each model, with error bars representing the 95% confidence intervals.*

## External Validation

Each model's performance on the internal manual test set was compared to external validation using the CT-RATE dataset. The CT-RATE study provided pseudo-labels from CT reports by using their fine-tuned RadBERT. Only four disease classes matched those in our study: atelectasis, lung nodules, emphysema, and pleural effusion. When restricted to the four overlapping classes, each model showed an increase in F1 performance when evaluated using the CT-RATE pseudo-labels. The model with the largest

F1 change was Llama-UM 0.67 [95% CI: 0.63, 0.71] (Internal) to 0.93 [95% CI: 0.92, 0.94] (CT-RATE) followed by RadBERT 0.77 [95% CI: 0.73, 0.80] (Internal) to 0.93 [95% CI: 0.92, 0.95] (CT-RATE).

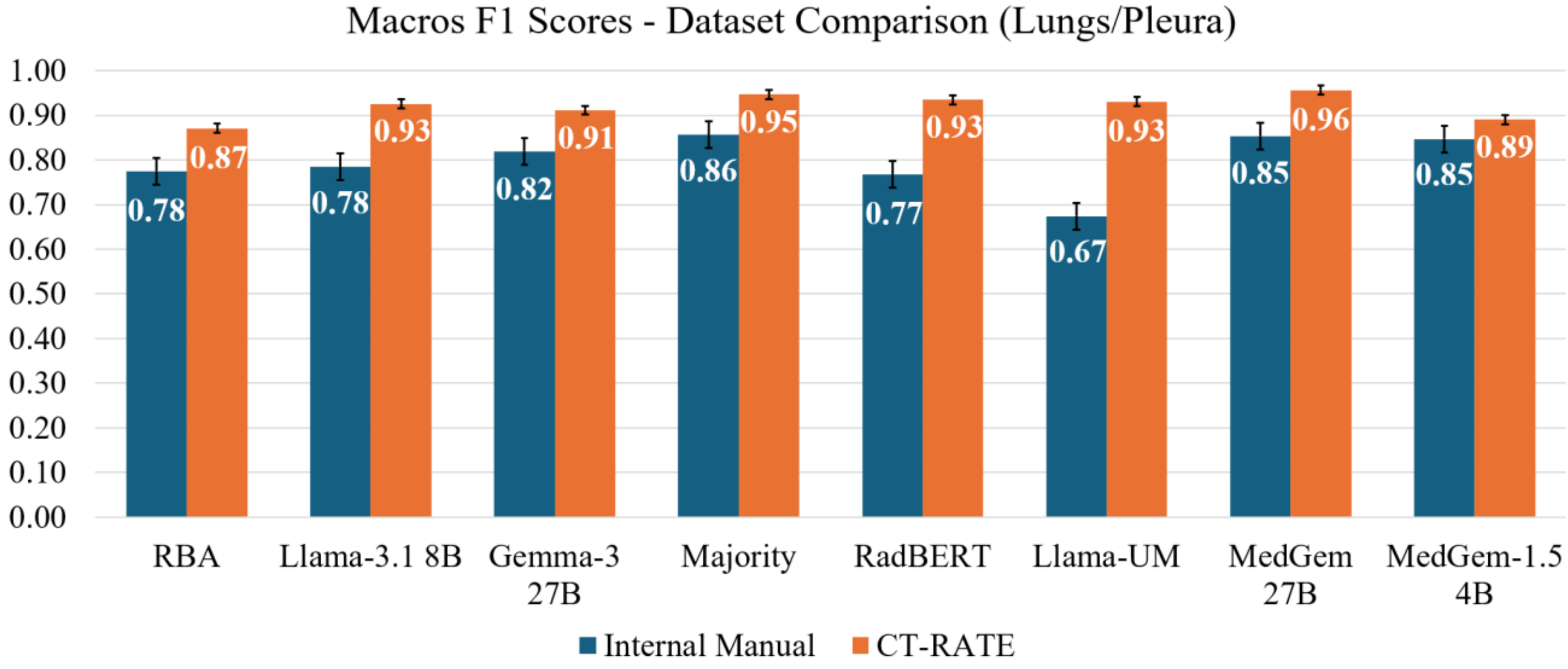

*Figure 4: Bar chart showing the macro-F1 scores for each model on both the internal manual set, and the CT-RATE set. The macro-F1 scores represent the average of the F1 scores for Atelectasis, Lung Nodules, Emphysema, and Pleural Effusion. The error bars represent 95% confidence intervals.*

## Sources of Label Disagreement

We reviewed the manual set to investigate why the labels for kidney lesion and atelectasis had the lowest average performance across all models. Those reports contained recurring phrases that produced disagreement between manual annotations and model-generated labels. The manual dataset was intended to reflect findings that are clinically actionable. For example, atelectasis was labeled positive, except “gravity-dependent” atelectasis was labeled as negative because it is a transient, non-concerning condition that can occur just from recumbency during the CT exam. Similarly, lesions “too small to characterize” were labeled as negative, as the radiologist could not confidently determine their nature. To capture this ambiguity, we introduced a “subjective” category to identify reports whose wording could reasonably lead to different binary decisions depending on the intended use of the labels. Reports containing kidney lesion, atelectasis, and the normal labels frequently contained subjective text: 96 of 343 reports for kidney lesion reports (28%) and 66 of 343 for atelectasis (19%). Examples of subjective texts and human versus model labeling disagreements can be seen in Figure 5.

| Label | Classification | Report Text | RBA | Llama-3.1 | Gemma-3 | Manual |
|---|---|---|---|---|---|---|
| Atelectasis | Definite Positive | "Scarring and **atelectasis** within the lungs." | 1✓ | 1✓ | 1✓ | **1** |
| Atelectasis | Subjective | "Aside from minimal subsegmental **gravity-*dependent* atelectasis**, the lungs are clear." | 0 ✓ | 1 X | 1 X | **0** |
| Kidney Lesion | Definite Positive | "Hypo-enhancing bilateral **renal lesions** are unchanged." | 1✓ | 1✓ | 1✓ | **1** |
| Kidney Lesion | Subjective | "There are multiple sub-centimeter hypodense **lesion** in the **right kidney** which are ***too small to characterize***." | 1 X | 1 X | 1 X | **0** |

*Figure 5 Representative report excerpts for atelectasis and kidney lesion, grouped by finding. For each finding, a definite case is shown first, followed by a subjective case in which the wording could reasonably support either binary label. Predictions from the RBA, Llama-3.1 8B, and Gemma-3 27B are shown alongside the corresponding manual annotation. Phrasing that produced the ambiguity is shown in bold.*

Recomputing F1 scores against the simplified manual set changed performance differently across the two classes. For kidney lesion, ΔF1 was positive for all models with the RBA having the largest average increase (0.41 [95% CI: 0.32, 0.50]) and Llama-UM having the smallest average improvement (0.06 [95% CI: -0.04, 0.15]). For atelectasis, all five LLMs improved, with MedGemma 27B showing the largest increase (0.32 [95% CI: 0.26, 0.40]), whereas the RBA and RadBERT showed performance favoring the original manual set, with ΔF1 spanning -0.05 [95% CI: -0.12, 0.02] and -0.06 [95% CI: -0.14, 0.02], respectively.

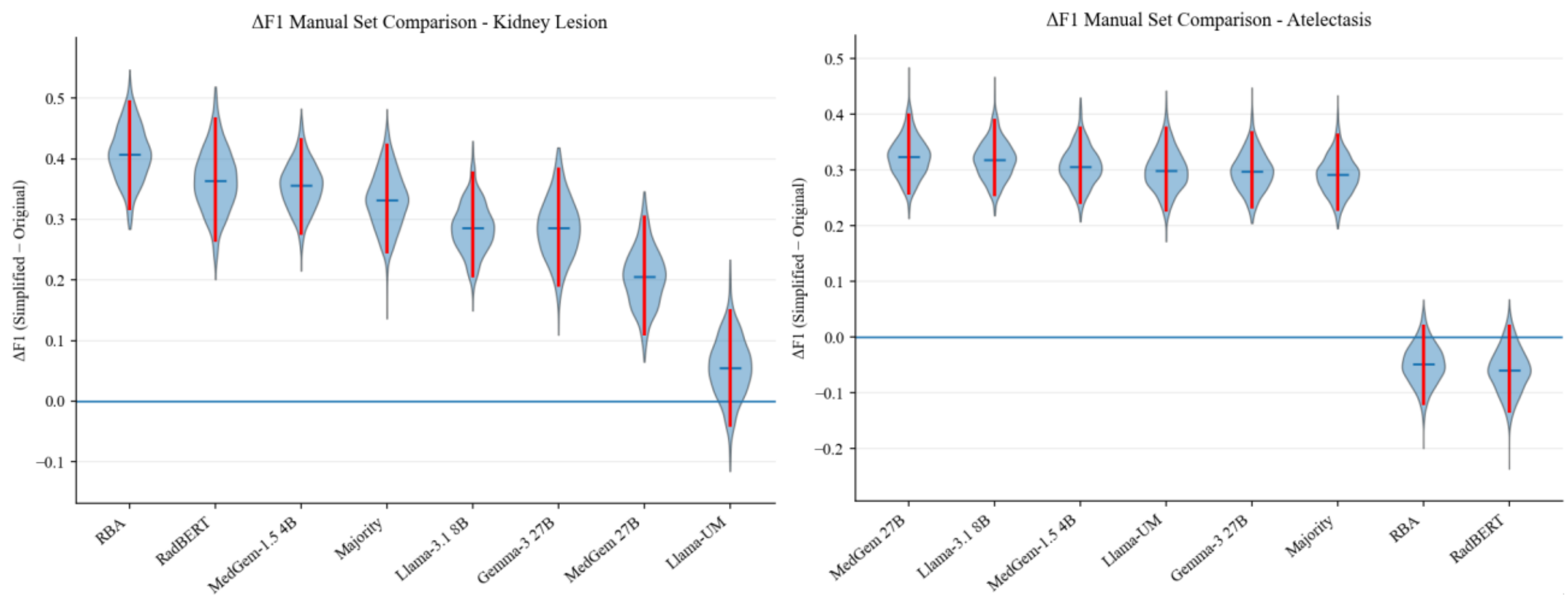

*Figure 6: Violin plots of F1 scores to compare each model across the originial and simplified internal manual test datasets. The width of the shape indicates the density of bootsrap samples with a particular value. The vertical red bar represents the 95% C.I. and the vetical bar is the mean value. A positive ΔF1 indicates a better performance on the simplified set and a negative ΔF1 indicates better performance on manual set.*

**Table 3: Average ΔF1 Values**

| Models | Kidney Lesion | Atelectasis |
|---|---|---|
| **RBA** | 0.41 [0.32,0.50] | -0.05 [-0.12,0.02] |
| **RadBERT** | 0.37 [0.26,0.47] | -0.06 [-0.14,0.02] |
| **Llama-3.1 8B** | 0.29 [0.20,0.38] | 0.32 [0.25,0.39] |
| **Llama-UM** | 0.06 [-0.04,0.15] | 0.30 [0.23,0.38] |
| **Gemma-3 27B** | 0.29 [0.19,0.39] | 0.30 [0.23,0.37] |
| **MedGemma 27B** | 0.21 [0.11,0.31] | 0.32 [0.26,0.40] |
| **MedGemma-1.5 4B** | 0.36 [0.27,0.43] | 0.31 [0.24,0.38] |
| **Majority** | 0.33 [0.24,0.43] | 0.29 [0.23,0.37] |

*Table 3: Average ΔF1 scores with 95% confidence intervals in brackets.*

# Discussion

In this retrospective study, five lightweight open-weight LLMs outperformed both a rule-based algorithm and a fine-tuned RadBERT model, for zero-shot multi-disease labeling of CAP CT reports across three organ systems. A majority-vote ensemble exceeded every individual model, suggesting that models with complementary error profiles can be combined to improve robustness.

Patterns of pairwise agreement varied with model types. The two MedGemma models, which share a training lineage, agreed more closely than any other pair. Llama-3.1 8B and Gemma-3 27B agreed marginally less but more consistently across classes despite coming from different families, a convergence that may reflect their common instruction-tuning rather than shared origin. The RBA and RadBERT also agreed closely, as expected, since RadBERT learned from RBA pseudo-labels. This dependency reached into performance: both shared the same worst-performing class and achieved near-identical macro F1 scores, indicating that RadBERT reproduced rather than corrected the RBA's failure modes. Llama-UM, by contrast, diverged from Llama-3.1 8B despite sharing a base architecture, likely due to its fine-tuning on conversational question-answer data rather than structured extraction.

Hierarchical clustering split the labels into disease and normal classes at the first bifurcation, with the normal classes drawing the lowest agreement across all models. Differing operational definitions account for this: the RBA demanded both exclusion of all predefined diseases and an explicit statement of normality, whereas the LLMs inferred normality without explicit wording or flagged abnormalities beyond the RBA's set. Asserting absence proved harder than detecting presence for every method.

Our findings extend prior work in two directions. D'Anniballe et al. (16) annotated CAP reports across the same three organ systems but built a dictionary for each label and retrained for each new label set; we achieved better performance with zero-shot prompting alone. Nowak et al. (9) found that open-weight LLMs outperformed rule-based and BERT-based extraction on chest radiograph reports, with Llama-3.1 8B reaching a macro-averaged F1 of 83.5%, closely matching what we observed for the same model on longer, anatomically broader CAP reports. In abdominal CT, LEAVS (15) reached 0.89 using chain-of-thought prompting within a tree-based decision structure; although the labeling schemes do not compare directly, unmodified zero-shot prompts recovered much of this accuracy at the cost of the richer label types such pipelines support.

Two analyses indicate that label definitions, rather than model capability, drove much of the residual disagreement. Every model agreed more closely with external CT-RATE labels than with our manual annotations, including the two that performed worst internally. A fine-tuned RadBERT generated those labels rather than a supervising radiologist, so they reflect a more literal reading of report text; dependent atelectasis counts as positive there and negative in our set. Recomputing scores against a simplified standard that disregarded clinical actionability produced a directional result: performance rose for every model on kidney lesion, but on atelectasis rose only for the LLMs and fell for the RBA and RadBERT. The RBA's rules explain this reversal, since they explicitly exclude dependent atelectasis and encode the same criterion our annotators applied, a behavior RadBERT inherited through its pseudo-labels. The direction of change therefore identifies which convention each method encodes. Because we derived the simplified labels from reviewing model disagreements rather than specifying them in advance, this comparison serves as a sensitivity analysis.

This distinction matters for how investigators validate automated labelers: a model that agrees closely with machine-generated labels may perform substantially worse against a clinically grounded standard. Neither convention is incorrect. A dataset assembled to train a detection model may reasonably record every mention of atelectasis, whereas one intended to flag patients for clinical review should not. Because LLMs can follow either definition when prompted but cannot infer which one an application requires, investigators should state annotation criteria explicitly.

Our study had limitations. We used zero-shot prompting to assess out-of-the-box performance, which likely underestimates what chain-of-thought prompting or purpose-built inference structures can achieve (15), and we did not fine-tune any of the LLMs, although instruction-based fine-tuning improves performance on comparable tasks (29). RadBERT learned from RBA pseudo-labels, limiting its independence as a comparator. Finally, we restricted analysis to binary labels to maintain comparability across methods; future work will move toward systems that represent linguistic ambiguity and clinical relevance, including multi-agent frameworks aligned to end-user specifications.

# Conclusion

In conclusion, lightweight open-weight LLMs outperformed both a rule-based algorithm and a fine-tuned RadBERT model for multi-disease annotation of CAP CT reports, generalizing across three anatomically distinct organ systems using zero-shot prompting alone and without task-specific development. Combining models with complementary strengths yielded further gains. Much of the remaining disagreement between models and human annotation traced to the interpretive criteria encoded in the labels rather than to model error, indicating that flexible labeling systems aligned to specific downstream applications may prove more valuable than further gains against any single fixed reference standard.

## Acknowledgements

Claude Opus 5 (Anthropic) was used to assist with editing manuscript wording and structure. It was not used to generate study data, perform analyses, or interpret results. All content was reviewed and verified by the authors, who take full responsibility for the final manuscript.

This study was supported in part by (Redacted).